\documentclass[10pt,twocolumn,letterpaper]{article}

\usepackage[pagenumbers]{cvpr} % To force page numbers, e.g. for an arXiv version

\usepackage{graphicx}
\usepackage{amsmath}
\usepackage{amssymb}
\usepackage{booktabs}

\usepackage[pagebackref,breaklinks,colorlinks]{hyperref}

\usepackage[capitalize]{cleveref}
\crefname{section}{Sec.}{Secs.}
\Crefname{section}{Section}{Sections}
\Crefname{table}{Table}{Tables}
\crefname{table}{Tab.}{Tabs.}

\def\confName{CVPR}
\def\confYear{2023}

\usepackage{latexsym}
\usepackage{comment}

\usepackage{color} % 最終的にはNG
\usepackage{multirow} %ok
\usepackage{algorithm, amsmath, amsfonts, amsthm, mleftright, multirow, bm, amssymb}

\usepackage[inline]{enumitem}
\usepackage{booktabs}
\usepackage{algpseudocode}
\definecolor{babypink}{rgb}{0.96, 0.76, 0.76}
\definecolor{green-yellow}{rgb}{0.68, 1.0, 0.18}
\definecolor{pastelorange}{rgb}{1.0, 0.7, 0.28}
\definecolor{paleblue}{rgb}{0.69, 0.93, 0.93}

\theoremstyle{definition}

\newcommand{\kosuke}[1]{\textcolor{black}{#1}}

\date{}

\begin{document}
%\linenumbers % camera readyで消す
\title{Robust Text-driven Image Editing Method that Adaptively Explores Directions in Latent Spaces of StyleGAN and CLIP}

\author{Tsuyoshi Baba\thanks{Equal contribution. This work was done when Tsuyoshi Baba was doing internship at NTT.}\;$^\dag$ \quad 
Kosuke Nishida\footnotemark[1]\;$^\ddag$ \quad Kyosuke Nishida$^\ddag$\\
$^\dag$Police Info-communications Research Center, National Police Academy \\
$^\ddag$NTT Human Informatics Laboratories, NTT Corporation\\
%1-1 Hikarinooka Yokosuka, Kanagawa, Japan \\
{\tt\small \{tsuyoshi.baba.um, kosuke.nishida.ap, kyosuke.nishida.rx\}@hco.ntt.co.jp}
}

\maketitle

\begin{abstract}
Automatic image editing has great demands because of its numerous applications, and the use of natural language instructions is essential to achieving flexible and intuitive editing as the user imagines.
%To solve this text-driven image editing task, we explore the latent semantics of the multi-modal feature space learned with contrastive language-image pre-training (CLIP) model.
%use contrastive language-image pre-training (CLIP) model that embeds visual and linguistic concepts into the multi-modal feature space.
%Our model performs only with the input of any natural language instruction and a strength hyperparameter, although previous work requires additional inputs that users must tune: such as a neutral text, a reference image, and hyperparameters.
A pioneering work in text-driven image editing, StyleCLIP, finds an edit direction in the CLIP space and then edits the image by mapping the direction to the StyleGAN space. At the same time, it is difficult to %choose appropriate hyper-parameters 
tune appropriate inputs
other than the original image and text instructions for image editing.
%To accomplish the editing with the minimal input, we construct the edit direction adaptively in StyleGAN and CLIP spaces with SVM and a gaussian mixture model (GMM). 
%interpret the latent semantics of StyleGAN and CLIP spaces with SVM and a gaussian mixture model (GMM).
In this study, we propose a method to construct the edit direction adaptively in \kosuke{the StyleGAN and CLIP spaces with SVM\@.} % and a gaussian mixture model (GMM).
Our model represents the edit direction as a normal vector in the CLIP space 
%learned with SVM 
obtained by training a SVM to classify positive and negative images. The images are retrieved from a large-scale image corpus, originally used for pre-training StyleGAN, according to the CLIP similarity between the images and the text instruction.
%positive and negative images that are retrieved on-the-fly according to the CLIP similarity to the instruction. 
%Second, our model decides the channels of the input image to be edited in the StyleGAN space %maps the vector to the StyleGAN feature space 
%by utilizing the distribution of the images estimated with a GMM.
We confirmed that our model performed as well as the StyleCLIP baseline, whereas it allows simple inputs without increasing the computational time.
\end{abstract}

\section{Introduction}
Automatic image editing has large demands %throughout industry 
because of its numerous applications.
%Generative Adversarial Networks (GANs) \cite{gan} have seen significant progress, and especially StyleGAN~\cite{StyleGAN} has achieved the success in the synthesis of high-resolution images because of its disentangled feature space~\cite{}.
%In order to semantically control the synthesized images, previous work utilized hoge\cite{} and fuga\cite{}.
Generative adversarial networks (GANs) \cite{gan} are one of the most common ways for synthesizing high-resolution images 
%Especially, StyleGAN~\cite{StyleGAN} has the semantically disentangled feature space~\cite{}, and thus it is widely used for the automatic image editing by 
and has been widely used for automatic image editing by 
using pre-defined attributes~\cite{attr1,attr2} and reference images~\cite{refimage1,refimage2}.
%geometryを入れる人もいる

\begin{figure}[t]
    \centering
    \includegraphics[width=0.48\textwidth]{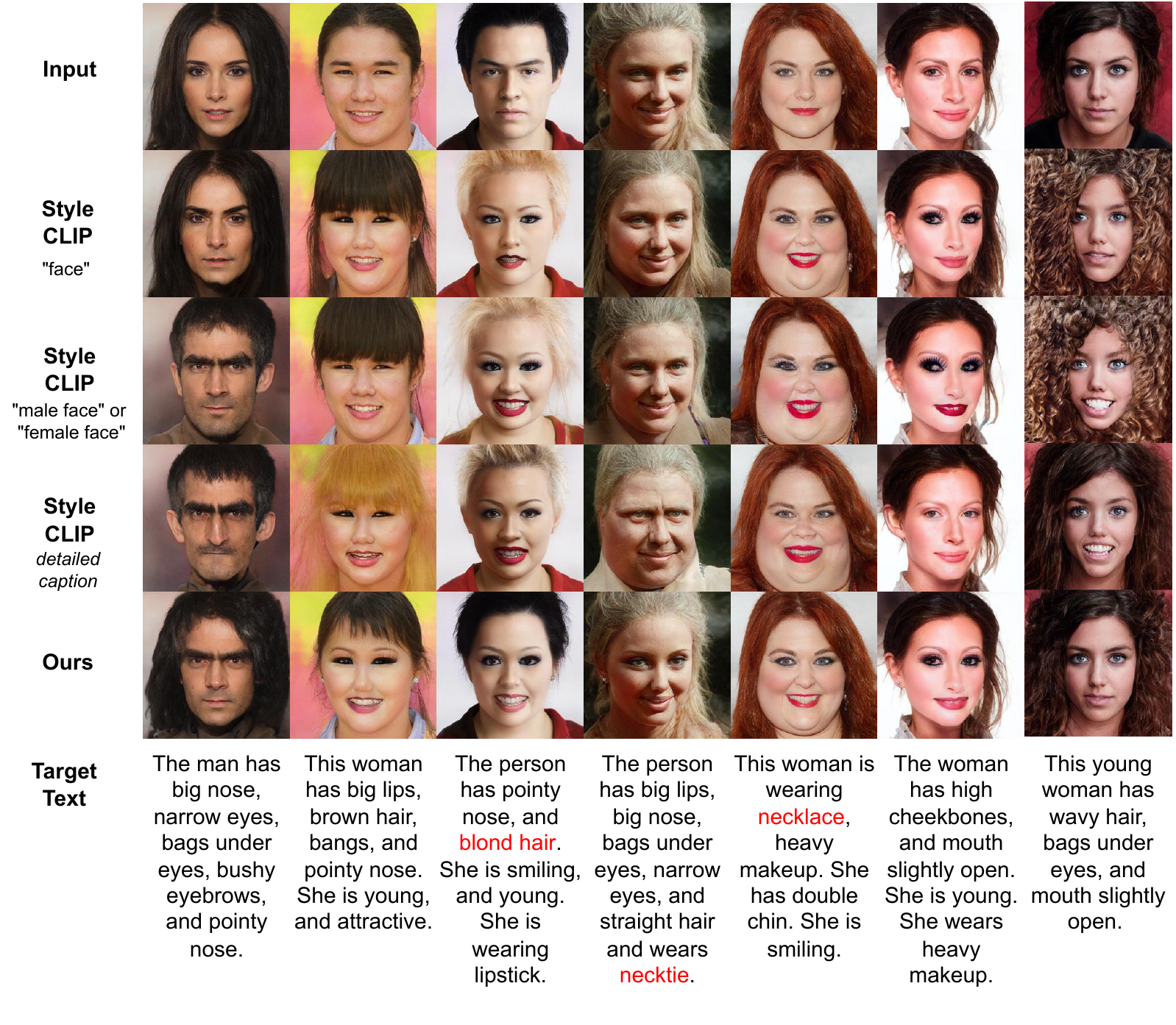}
    \caption{Examples of images edited using StyleCLIP-FEU and StyleCLIP.
    The neutral text $t_0$ of StyleCLP is (a)``face'', (b)``male face'' or ``female face'', and (c) detailed caption. StyleCLIP-FEU edits more attributes in the instruction without $t_0$. The apparent failures of StyleCLIP-FEU are colored in red.}
    \label{fig:top}
\end{figure}

Image editing using natural language instructions
is important for developing easy-to-use applications that enable flexible and intuitive editing as imagined by the user.
This challenging task involves multi-modal understanding of images and natural language. 
Here, the contrastive language-image pre-training (CLIP) model embeds both visual and linguistic concepts into the multi-modal feature space~\cite{CLIP}. Therefore, CLIP provides an opportunity to explore methods of performing this task.

%StyleCLIPの導入．課題も簡単に
The pioneering work StyleCLIP~\cite{StyleCLIP} adds a vector to the original image representation in the StyleGAN feature space, which is known to be well disentangled~\cite{disentangle1, disentangle2, disentangle3, disentangle4}.
%これ関連研究でいい気がするけど，sparseに言及するならあった方がいい
It proposed two types of method: The first type, consisting of latent optimization and latent mapper, computes the back-propagation through a neural network by maximizing the CLIP similarity between the output image and the given natural language instruction. 
Therefore, these methods have a large computational cost for each user's input.

The second type, called global direction, obtains an edit direction in the CLIP space and then maps it to the StyleGAN feature space.
The edit direction is defined as a vector from the text description of the original image to the instruction by using the CLIP text encoder. %The edit direction is mapped with a sparsity hyperparameter to affect only the attributes mentioned in the instruction. 
Thus, the user must manually specify the original image description. % and the sparsity hyperparameter.
%they requires that the user manually describes the original image to obtain the origin of the edit direction with CLIP text encoder and spacifies the sparsity hyperparmeter in the StyleGAN feature space. 
%However, to obtain the edit direction, StyleCLIP requires additional input that the user must tune.

%Therefore, we cannot develop a fast and easy-to-use image editing application based on StyleCLIP.
%Although some papers extended StyleCLIP~\cite{mapper1,mapper2,interactive,anyface,clip2stylegan, ppe}, they did not focus on such problems except for \cite{stylemc} that followed the first method and accelarated the optimization.

%提案手法
%やっぱりこの時点でsparsity parameterとneutral textは触れておきたい
In order to make a fast and easy-to-use (FEU) image editing application, we propose StyleCLIP-FEU, which takes the second approach by interpreting the latent semantics of the StyleGAN and CLIP space.
Inspired by InterfaceGAN~\cite{InterfaceGAN}, to eliminate the manual description of the original image, we define the edit direction as a normal vector learned with SVM for classifying positive and negative images that are retrieved on-the-fly with CLIP.
%In addition, we adaptively determine the sparsity hyperparameter according to the distribution of the images in the feature space. %We estimate the distribution with a gaussian mixture model, and use the cluster-wise standard deviations to obtain the thresholds.
Although some papers have extended StyleCLIP~\cite{mapper1,mapper2,interactive,anyface,clip2stylegan, ppe}, they did not focus on the aforementioned problems, except for \cite{stylemc}, which followed the first method and accelerated the optimization, and \cite{StyleGAN-T}, which followed the second method and added further pre-training to align the StyleGAN space and the CLIP space.
%We use gaussian mixture model to model the distribution.

\kosuke{
Experimental results showed that StyleCLIP-FEU edits facial images naturally under guidance of natural language instructions. Moreover, we observed that StyleCLIP is influenced by the original image description.
}

\iffalse
%評価
%FIDやcos simの安定性を検証したい
%single attributeとmulti attribute
Our main contributions are as follows.
\begin{itemize}
\item To achieve flexible image editing with intuitive natural language instruction, we propose StyleCLIP-FEU that performs only with the input of the instruction and a strength hyperparameter.
\item We conducted extensive experiments to investigate the image editing performance of methods that produce the results within a minute. The experiments shows ...
\item We found that ...
\end{itemize}
\fi

\section{Preliminaries}
\subsection{Task Definition}
We define the image editing with natural language instructions %as following:
%\begin{defi}[Image Editing with Natural Language Instruction]
as a task where
a model generates an image $x'$ based on an original image $x$ by following an natural language instruction $t$.
%\end{defi}
Additionally, we assume that an image corpus (without any annotations) $I$ is available.

\subsection{CLIP}
CLIP~\cite{CLIP} embeds images and texts that mention similar concepts close to each other in a multi-modal feature space $M$.
CLIP has an image encoder $\mathrm{CLIP_{image}}$ and a text encoder $\mathrm{CLIP_{text}}$. It calculates an image embedding $\bm{h}_{\mathrm{image}}$ and text embedding $\bm{h}_{\mathrm{text}}$ with each encoder.
CLIP was trained with 400M pairs of images and their captions collected from websites.
It was trained to increase $\cos (\bm{h}_{\mathrm{image}}, \bm{h}_{\mathrm{text}})$ between positive pairs more than in-batch negative pairs.

\subsection{StyleGAN}
StyleGAN~\cite{StyleGAN,StyleGAN2} generates style-controlled images because of its disentangled feature space.
It is composed of a mapping module and a synthesis module.
Each module consists of a $N_{\mathrm{map}}$ and $N_{\mathrm{syn}}$ layer neural network.
StyleGAN first obtains a latent feature $\bm{w} \in W $ from Gaussian noise $\bm{z}$ with the mapping module.
Then, it affinely transforms $\bm{w}$ into $\bm{s}_l ~(l =1, \cdots,N_{\mathrm{syn}})$, which are fed to the $l$-th layer of the synthesis module. 
The concatenation of $\bm{s}_l$ over all layers, $\bm{s} \in S$, is called a style feature.

\subsection{StyleCLIP}
\cite{StyleCLIP} developed StyleCLIP by combining StyleGAN and CLIP for image editing with natural language instructions. 
StyleCLIP consists of three methods: latent optimization, latent mapper, and global directions.
Among them, we focused on the global directions method because the other methods require the optimization of the neural network for each $t$, and the authors reported that the features in the global directions are most disentangled.

\cite{StyleCLIP} formulated an edit as the direction in the CLIP space $M$ from the neutral text (the text describing the original image, \textit{e.g.}, ``face with hair'') $t_0$ to the target text (the instruction, ``face with blond hair'') $t$. Thus, StyleCLIP requires neutral text in addition to the inputs defined in our task definition.
The model maps the edit direction in $M$ to a vector $\Delta \bm{s} \in S$.
%to a vector $\Delta \bm{s}$ 
%the difference between the neutral text (the text describing the original image) $t_0$ and the target text (the instruction) $t$ in the sto the style space $S$.
Then, the synthesis module $G$ generates an edited image $x' =G(\bm{s} + \alpha \Delta \bm{s})$, where $\bm{s}$ is the style feature of the original image $x$ encoded with an image encoder such as E4E~\cite{e4e} and $\alpha$ is a hyperparameter.
%Here, $\alpha$ is a strength hyperparameter.

$\Delta s_c$ represents to what extent the edit direction in $M$ is related to the attributes corresponding to the $c$-th channel in the disentangled style space $S$.
%the strength of the relationship between the $c$-th channel in the disentangled style space $S$ and the edit direction. 
Thus, StyleCLIP defines $\Delta s_c$ on the basis of the dot product of two vectors in $M$, $\Delta \bm{i}_c$ and $\Delta \bm{t}$.
Here, $\Delta \bm{i}_c$ is a vector mapped from the channel $c$ direction in $S$.
$\Delta \bm{t}$ is a vector that indicates the edit.

%\paragraph{Direction corresponding to a channel}
\paragraph{Construction of $\Delta \bm{i}_c$}
First, StyleCLIP obtains the CLIP space direction $\Delta \bm{i}_c$ that corresponds to channel $c$ of the style space. 
From the image corpus $I$, StyleCLIP samples 100 images $x_i \quad (i =1,\cdots ,100)$ and obtains their style features $\bm{s}_i$ with an E4E encoder.
The model calculates the standard deviation $\bm{\sigma} \in \mathbb{R}^C$, where $C$ is the number of the channels.
The model perturbs the channel $c$ of the image feature $\bm{s}_i$ and generates pairs of images:
%$x_{i,c}^{\pm} =G(\bm{s} \pm 5 \sigma_c \bm{e}_c)$,
\begin{equation}
\label{eq:x_ic}
x_{i,c}^{\pm} =G(\bm{s}_i \pm 5 \sigma_c \bm{e}_c),
\end{equation}
where $\bm{e}_c \in S$ is a unit vector in channel $c$.
The model obtains the direction in the CLIP space $M$
\begin{equation}
\label{eq:delta_i_c}
%\[
\Delta \bm{i}_c = \mathrm{Ave}_i \left( \mathrm{CLIP_{image}}(x_{i,c}^+) - \mathrm{CLIP_{image}}(x_{i,c}^-) \right).
%\]
\end{equation}

%\paragraph{Direction corresponding to the editing}
\paragraph{Construction of $\Delta \bm{t}$}
Then, StyleCLIP obtains the CLIP space direction $\Delta \bm{t}$ that corresponds to the edit.
In StyleCLIP, the edit is defined with the target text $t$ and the neutral text $t_0$. Also, the model prepares 80 prompts $p$ (\textit{e.g.}, ``A photo of'') and concatenates each of them with the text, which is denoted by $[ \cdot ; \cdot ]$. The model obtains the direction in the CLIP space $M$
\begin{equation}
\label{eq:delta_t}
\Delta \bm{t} = \mathrm{Ave}_p \left( \mathrm{CLIP_{text}}([p; t]) - \mathrm{CLIP_{text}}([p; t_0]) \right).
\end{equation}

\paragraph{Construction of $\Delta \bm{s}$}
It is expected that $\Delta s$ affects only the attributes mentioned in the target text.
Therefore, instead of the simple dot product of $\Delta \bm{i}_c$ and $\Delta \bm{t}$, StyleCLIP introduces a hyperparameter $\beta$ to make $\Delta \bm{s}$ sparse.
Also, the model normalizes $\Delta \bm{s}$ so that its maximum value is 1:
%\[
\begin{equation}
\label{eq:delta_s}
\Delta s_{0,c} = \left\{ \begin{array}{cl} \Delta \bm{i}_c \cdot \Delta \bm{t} & \mathrm{if}~ |\Delta \bm{i}_c \cdot \Delta \bm{t}|\geq \beta \\
0 & \mathrm{otherwise} \end{array} \right.,
\end{equation}
%\]
\begin{equation}
\label{eq:norm}
%\[
\Delta \bm{s} =\frac{1}{\max_c (s_{0,c})} \Delta \bm{s}_0.
%\]
\end{equation}

\paragraph{Limitations} StyleCLIP requires the neutral text $t_0$ as an additional input. $t_0$ requires the user to describe the original image in terms of the attributes mentioned in the instruction by trial and error. 
In essence, what attributes are edited, or not edited, can be derived from the instruction $t$ and the original image $x$.
For the efficient usage of an editing system, it is desirable that the model does not depend on such redundant input.

\section{Proposed Method}
\begin{figure}[t]
    \centering
    \includegraphics[width=0.32\textwidth]{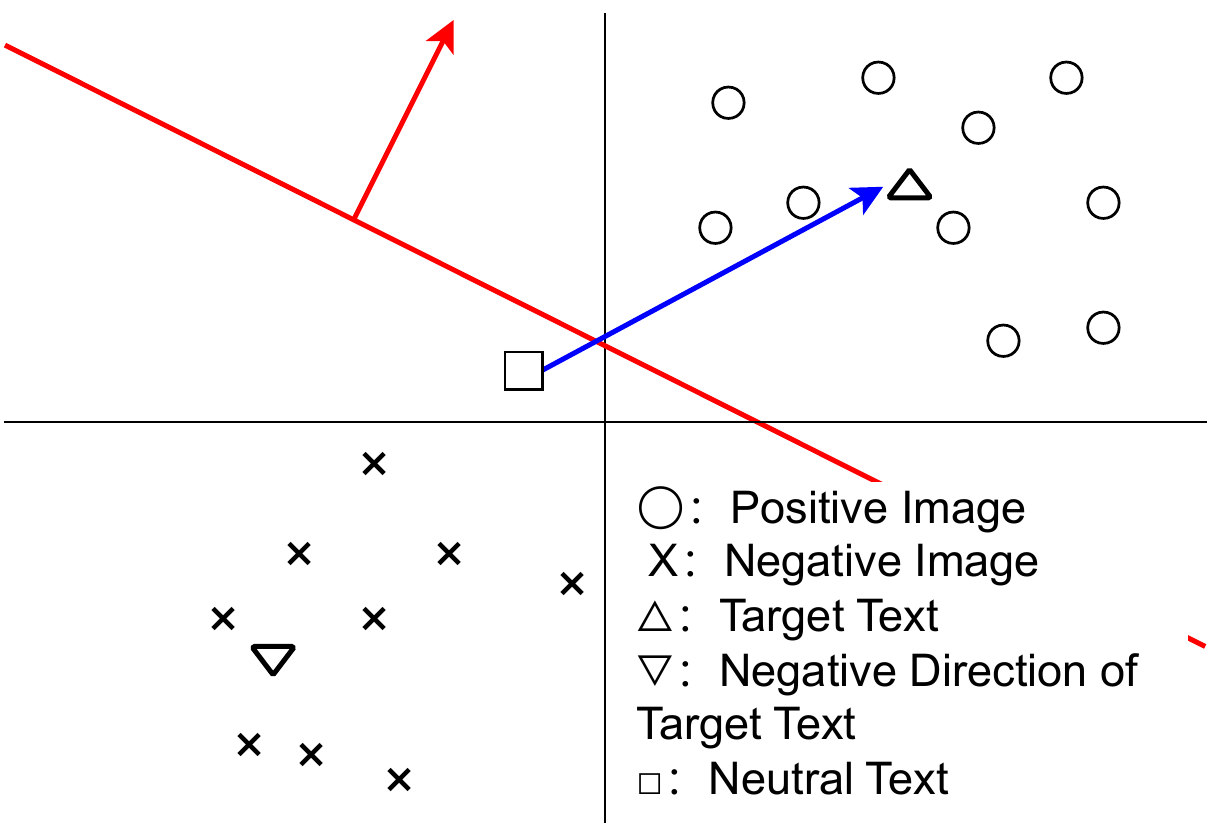}
    \caption{
    Construction of $\Delta \bm{t}$ in StyleCLIP (shown in blue) and the proposed method (red).
    StyleCLIP define $\Delta \bm{t}$ as the vector from $t_0$ to $t$.
    The proposed method use the normal vector of the hyperplane learned with SVM.}
    \label{fig:proposed}
\end{figure}

To eliminate the neutral text, we use a normal vector leaned with SVM for $\Delta \bm{t}$ instead of the edit direction from $t_0$ to $t$, as shown in Figure~\ref{fig:proposed}.
First, we retrieve 100 positive images $X^+$ and negative images $X^-$ from $I$ by referring to the cosine similarity with the text embedding of the instruction $\mathrm{CLIP_{text}} (t)$.
%This operation can be conducted within XXX seconds by using faiss library~\cite{FAISS}.
Then, we train a linear SVM for classifying the images in order to obtain the hyperplane $\hat{\bm{w}}^{\top} \bm{h}_{\mathrm{image}, i} +\hat{b}=0$ and its normal vector $\bm{n}=\hat{\bm{w}}$, where $\hat{\bm{w}}$ and $\hat{b}$ are learned parameters, and $\bm{h}_{\mathrm{image},i} =\textrm{CLIP}_{\textrm{image}}(x_i)$ where $x_i$ is an image in $X^{+}$ or $X^{-}$.
The SVM is formulated as follows:
%\[
\begin{equation}
%\nonumber
\label{eq:svm}
\underset{\bm{w}, b}{\textrm{min}}~~ \frac{1}{2}\| \bm{w} \|^2 + \sum_{x_i, y_i} \max\left(\left(1 - y_i(\bm{w}^{\top} \bm{h}_{\mathrm{image},i} + b)\right) , 0\right),
\end{equation}
%\]
where $y_i = \pm 1$ is a label of $x_i$. 
Finally, we define $\Delta \bm{t} =\bm{n}$.

%iterfaceganにここで言及
\kosuke{\cite{InterfaceGAN} proposed InterfaceGAN that trains a SVM in the style space for editing in terms of pre-defined attributes.} We use this technique to obtain the edit direction from any user's instruction on-the-fly.

\section{\kosuke{Experiments}}
We conducted experiments on image editing with natural language instruction in the face domain. 

\subsection{Dataset}
StyleCLIP-FEU requires no image-caption paired dataset for training.
We used Multi-Modal-CelebA-HQ as the validation set to select the hyperparameter and the test set. We also used FlickrFaces-HQ (FFHQ) as the image corpus $I$.

%\subsubsection{Multi-Modal-CelebA-HQ}
%\noindent \textbf{Multi-Modal-CelebA-HQ}~
Multi-Modal-CelebA-HQ \cite{TediGAN} contained ten automatically annotated captions for each image in the CelebA dataest~\cite{CelebA-HQ}.
We sampled 100 images and 10 captions randomly from the original training set and used them as the validation set. Moreover, we sampled 50 images and 10 captions randomly as the test set from the original test set and used them as the test set.
%We used the images and captions of the dataset for the input images and the instruction texts in our evaluation. We sampled 1000 captions from the test split for the instruction randomly. To avoid the overlap between the input image and the instruction, we sampled 1000 images from the training split for the input image randomly. Note that we used the captions as the instruction to evaluate the editing performance with natural language instruction, while previous work used attribute names as the instruction.

%\noindent \textbf{FlickrFaces-HQ (FFHQ)}~
FFHQ \cite{StyleGAN} has 70,000 face images and was used for pre-training StyleGAN.

%\noindent \textbf{Car}
%\subsubsection{Car}

\subsection{Implementation}
We used StyleGAN2~\footnote{https://github.com/NVlabs/stylegan2} and CLIP-ViT-B/32~\footnote{https://github.com/openai/CLIP} as the pre-trained models. 

We searched for the hyperparameter $\alpha \in [2.0, 6.0]$ by using step size of 0.5 and for $\beta \in [0.1, 0.2]$ by using a step size of 0.05 to maximize the cosine similarity between the edited image and the target text $t$ in the CLIP space. All methods selected $\alpha = 6.0$ and $\beta = 0.1$. The other hyperparameters followed the selections in  \cite{StyleCLIP}.

\subsection{Compared Models}
We used %two methods that can edit images with unseen instruction in a minute for comparison.
StyleCLIP global direction~\cite{StyleCLIP}. %, which synthesis high-quality image editing because of multi-modal CLIP space, although it requires the neutral text and the sparsity hyperparameter.
We evaluated three neutral texts: (a)``face'', (b)``male face'' or ``female face'', and (c)
a detailed description of the original image that was randomly sampled from the ten annotated captions in Multi-Modal-CelebA-HQ.

%\subsection{Metrics}
%We automatically evaluate the reality of the images with Fr\'{e}chet Inception Distance (FID)~\cite{fid}, and the accuracy of the editing to the instruction with CLIP cosine similatity.

%For the human evaluation, we asked three workers to ranked the images generated by StyleCLIP and StyleCLIP-FEU with two criteria: the accuracy to the instruction and the naturality of the image. To evaluate the accuracy, we automatically divided the instruction to the attributes, and the workers rank the images with regard to each attribute. We also calculate the overall accuracy by ranking the images according to the sum of the ranks of all attributes. We randomly sampled 100 image-text pairs for this evaluation.

%\subsection{Implementation}
%We retrieved 1000 positive and negative images, respectively, and use $\alpha =5.0$. For StyleCLIP, we tuned $\alpha =3.0$ and $\beta =0.10$ with preliminary experiment. We use `face' for the neutral text. For StyleMC, we followed their default implementation except for the strength hyperparameter $\alpha =10$.

\subsection{Results and Discussion}

\begin{figure}[t]
    \centering
    \includegraphics[width=0.48\textwidth]{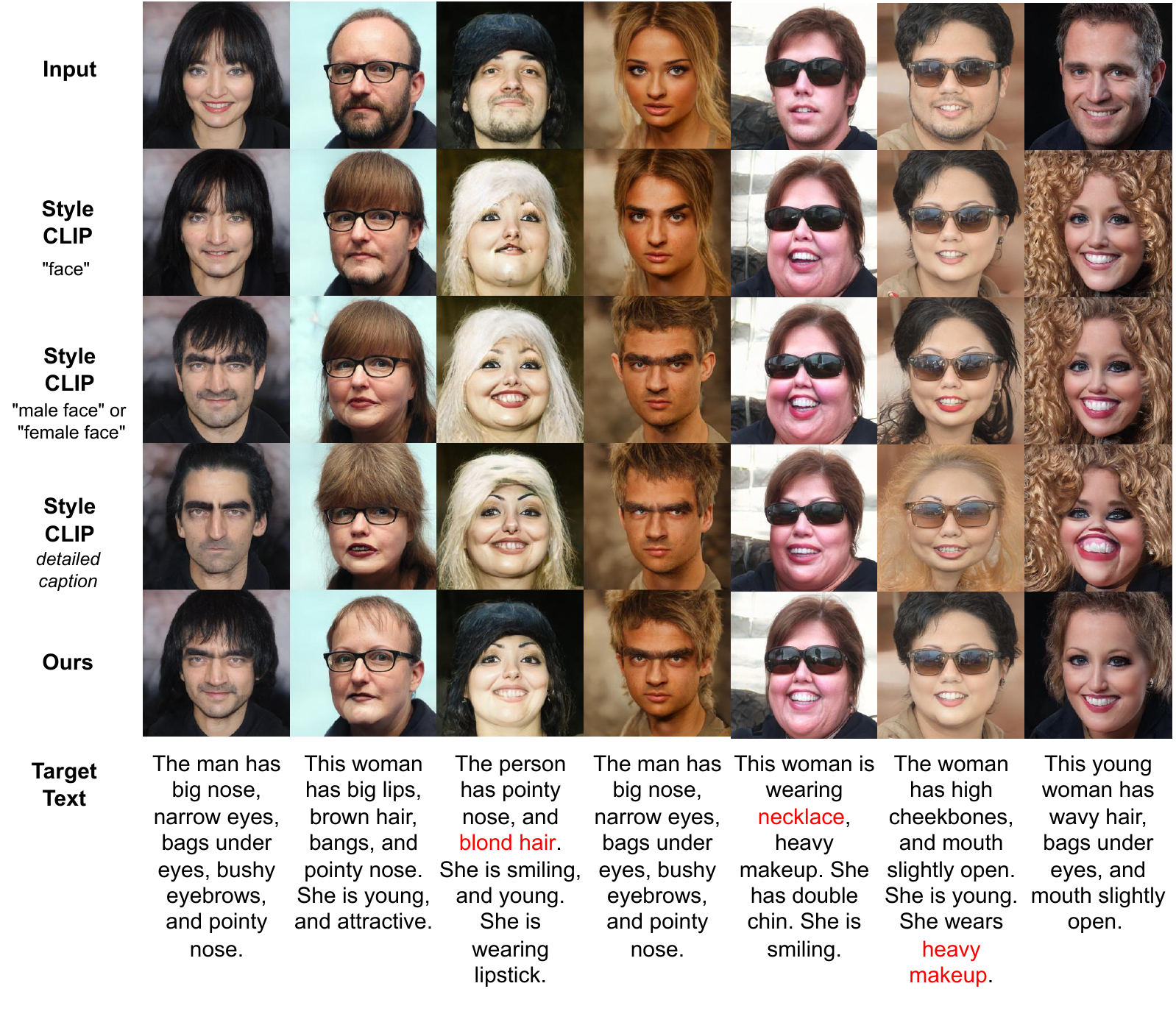}
    \caption{Examples of images edited using StyleCLIP-FEU and StyleCLIP in the setting where the gender of the original image is different from the gender specified in the instruction. The apparent failures of StyleCLIP-FEU are colored in red.}
    \label{fig:main}
\end{figure}

\paragraph{Sensitivity to neutral text.} As shown in Figure~\ref{fig:top}, the image generated by StyleCLIP depends on the neutral text. Especially, as shown in Figure~\ref{fig:main}, using ``male face'' or ``female face'' as $t_0$ improved the quality of the generation images if the gender of the original image is different from the gender specified in the instruction. StyleCLIP with the detailed caption successfully edited the images in some cases, but there were also cases in which the mouth or the shape of the face appeared unnatural. In comparison, StyleCLIP-FEU performed stable editing without $t_0$.

\begin{figure}[t]
    \centering
    \includegraphics[width=0.48\textwidth]{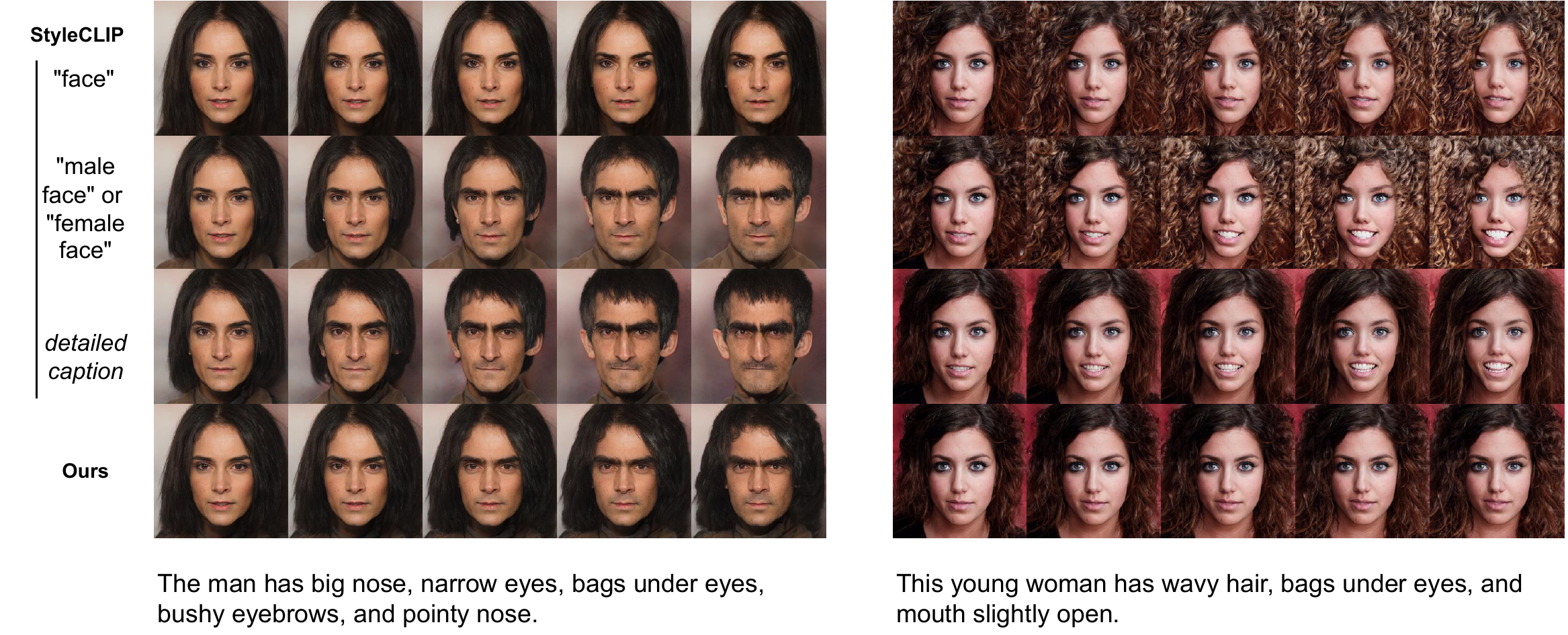}
    \caption{Examples of edited images when $\alpha$ is changed from 2.0 (left) to 6.0 (right). $\beta$ is 0.1.}
    \label{fig:alpha}
\end{figure}

\begin{figure}[t]
    \centering
    \includegraphics[width=0.48\textwidth]{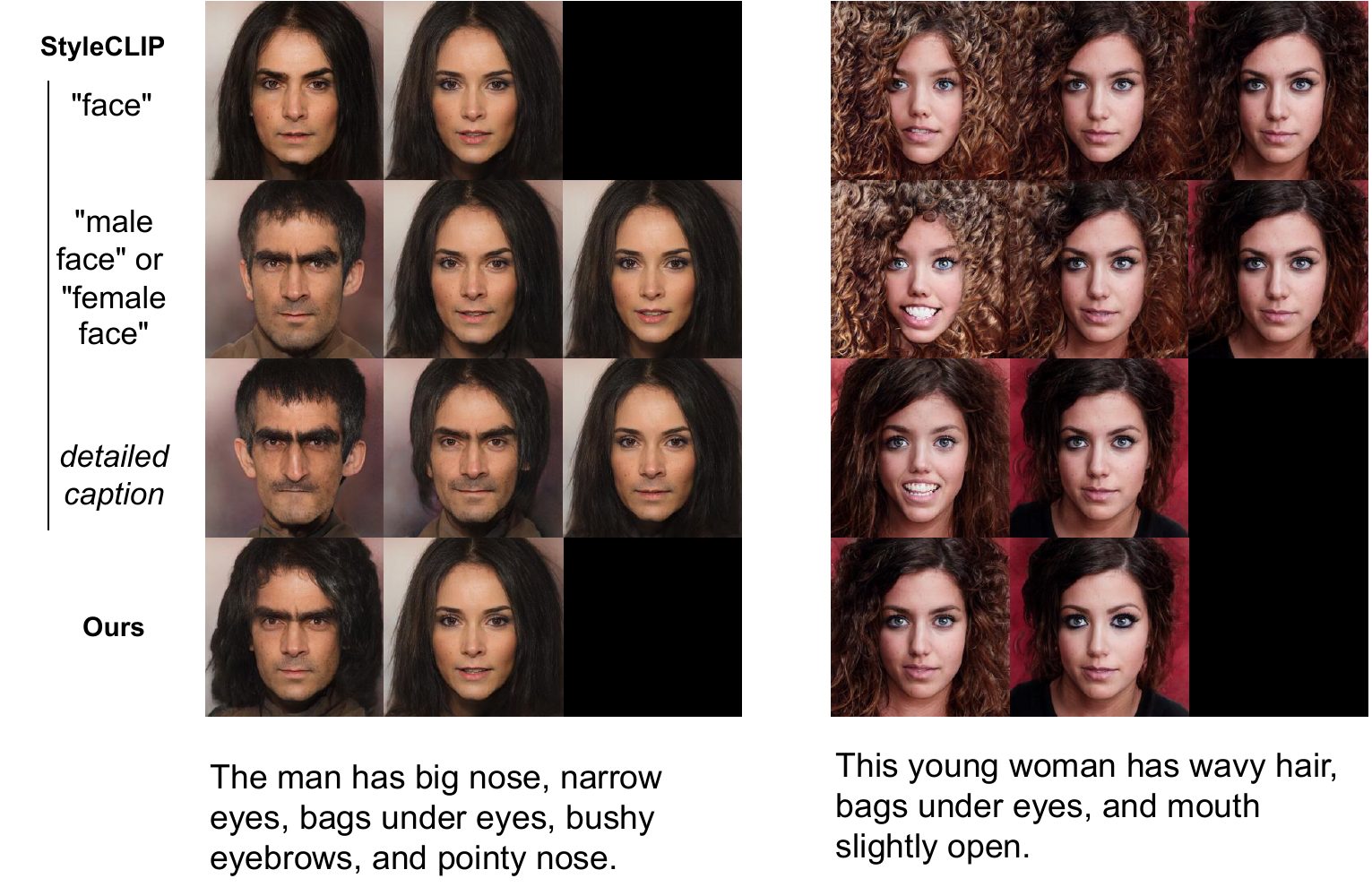}
    \caption{Examples of edited images when $\beta$ is changed from 0.1 (left) to 0.2 (right). $\alpha$ is 6.0.}
    \label{fig:beta}
\end{figure}

\paragraph{Sensitivity to hyperparameters.}
Figure~\ref{fig:alpha} shows examples of the effect of changing $\alpha$ from 2.0 to 6.0.
Compared with StyleCLIP-FEU, we observed that StyleCLIP caused excessive editing with large $\alpha$ and tuning $\alpha$ was difficult. 
Figure~\ref{fig:beta} shows examples of the effect of changing $\beta$ from 0.1 to 0.2. StyleCLIP and StyleCLIP-FEU with the large value of $\beta$ failed to generate images because of the divergence of normalization.
However, the small value of $\beta$ caused editing of attributes that were irrelevant to the instructions. Adaptive selection of $\beta$ will be part of our future work.
%すぎると操作対象属性の選定に関わる$\Delta s$がすべて0となり，正規化に問題が発生して生成に失敗してしまう例が多く見られた．

\begin{figure}[t]
    \centering
    \includegraphics[width=0.48\textwidth]{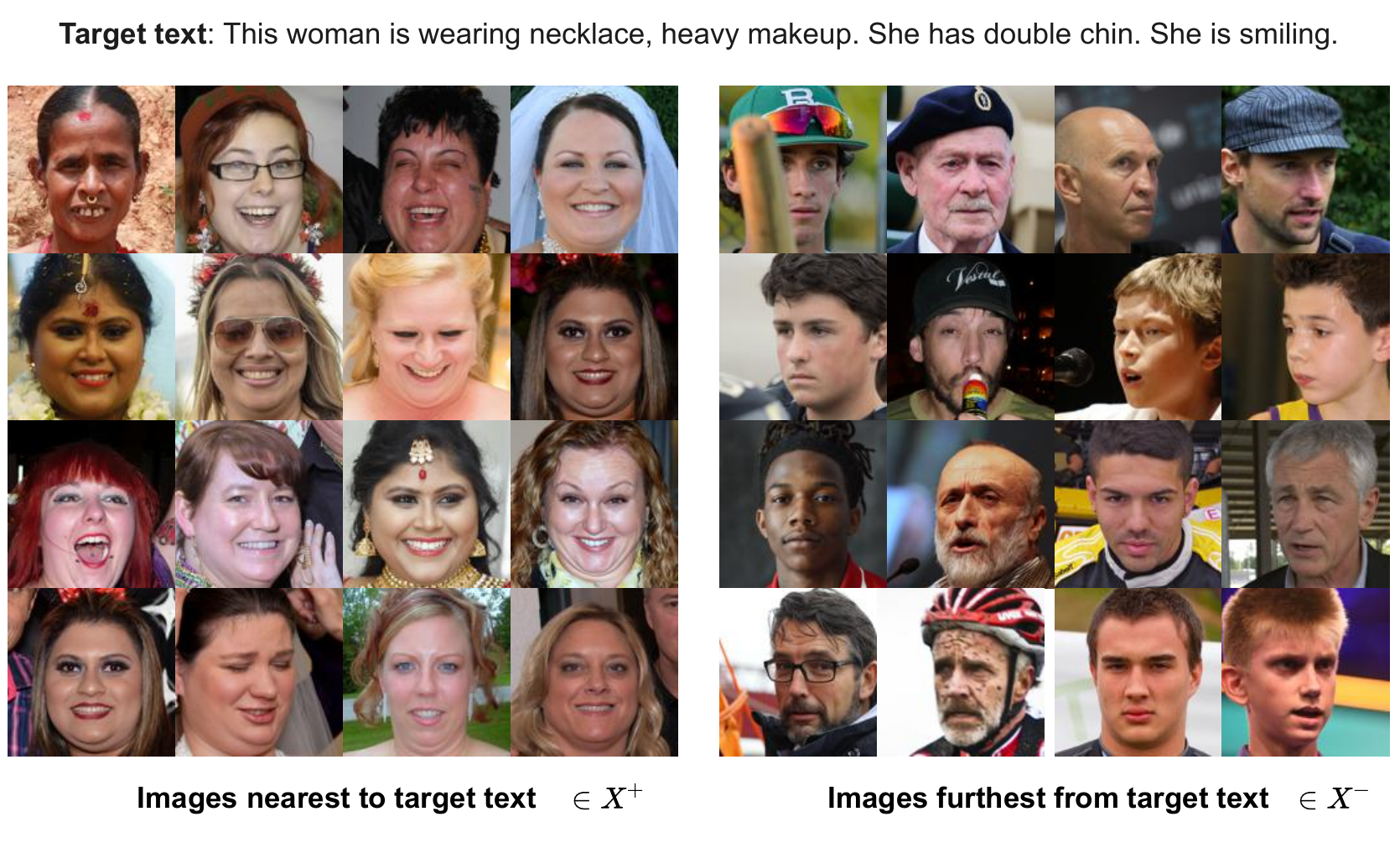}
    \caption{Top-16 positive images $X^{+}$ and negative images$X^{-}$.}
    \label{fig:pn}
\end{figure}

\paragraph{Positive and negative images that determine separating hyperplane.}
Figure~\ref{fig:pn} shows the positive and negative images that StyleCLIP-FEU retrieved in the CLIP space.
In this example, $X^+$ includes the images satisfying the ``women'', ``double chin'', and ``smailing'' instructions. 
%$t^\mathrm{target}$を用いて取得したCLIP空間における近傍画像セット$X^+$および遠方画像セット$X^{-}$を示す．この例では``woman''や``double chin''，``smiling''については編集意図に適合した画像が$X^+$に含まれているが、``necklace''が含まれる画像は殆ど無かったため，分離超平面が``necklace''を考慮できなかった．今後，さらなる編集精度改善のためには，CLIP空間における情報検索において，$t^\mathrm{target}$で指定した編集指示の一部の属性のみに関する画像が$X^+$を占有してしまう問題の解決が必要である．

%\textbf{計算時間について　}
\paragraph{Inference time.}
The editing of one image with one instruction took 0.37s with StyleCLIP on a NVIDIA Tesla T4 GPU. The difference between StyleCLIP and StyleCLIP-FEU is the image retrieval and SVM training, which took less than 0.01s.
%The editing of one image with one instruction took 3.67s with StyleCLIP and 33.4s with StyleCLIP-FEU.
%The bottleneck of StyleCLIP is the retrieval of the images, and we can accelarate it with FAISS~\cite{FAISS}.

\paragraph{Subjective Evaluation.}
We asked two workers to evaluate the images generated by StyleCLIP and StyleCLIP-FEU in terms of two binary criteria: the accuracy of gender editing and the naturalness of the image, such as the shape of the face and the hair. The ground-truth gender is defined by the instruction $t$ if $t$ mentions the gender. Otherwise, it is the same as the original image.  Cohen's $\kappa$ between the annotators was 0.92 and 0.63.

StyleCLIP-FEU outperformed StyleCLIP on gender and naturalness; thus we confirmed that StyleCLIP-FEU has the capability of following $t$ and generating robustly.
In addition tuning of the neutral text is difficult because the detailed description degraded the naturalness of the images.

\begin{table}[t]	
\centering
\small
  \caption{Results of subjective evaluation.}%性別の正解率および顔が崩れていない割合．}
  \label{table:sub}
  \vspace{.5em}
  \begin{tabular}{c|c|c|c}
  \toprule
  %\hline
          Model & $t_0$  & Gender & Natural \\ \toprule
          %\hline 
    &  ``face'' & 86.6 & 82.0 \\ 
    StyleCLIP & ``male face'' or ``female face'' & 87.1 & 76.9 \\ 
    & detailed caption & 80.8 & 66.0 \\ \midrule
    Ours & -- & \textbf{87.5} & \textbf{95.7} \\ %\hline
    \bottomrule
  \end{tabular}
\end{table}

%手法が5つ
%StyleCLIP
%proposed
%proposed w/ GMM
%proposed w/ beta
%proposed w/ SVM

%自動評価指標
%MP: 元画像との類似度と，テキストとの類似度を取る指標 maniGAN https://openaccess.thecvf.com/content_CVPR_2020/papers/Li_ManiGAN_Text-Guided_Image_Manipulation_CVPR_2020_paper.pdf
% (1-diff)xsim, diffがpixelレベル，sim=CLIP

\section{Related Work}
%CLIP以前はみたいな書き方に
%StyleCLIP参照
Some studies had tackled image editing with natural language instructions before the publication of CLIP~\cite{sisgan,tagan,manigan}. They assumed that a text-image paired corpus was available to acquire a multi-modal understanding.

%StyleCLIP
StyleCLIP proposed two ways of combining StyleGAN and CLIP.
The first way involves optimization with a neural network and requires a large computational cost.
\cite{mapper1}, \cite{mapper2}, and \cite{ppe} improved the quality of the output.
The second way is a fast algorithm that uses the mapping of the edit direction. \cite{interactive} tackled the interactive image editing task with this sort of method. However, methods of this type requires an additional input. 
\kosuke{
\cite{StyleGAN-T} extended this type; they additionally trained StyleGAN to align the style space and the CLIP space, which took five weeks on 64 NVIDIA A100 GPUs. By comparison, StyleCLIP-FEU does not require any additional training besides the lightweight SVM training.
}

%ここにhairclipとかstyletransfer系を入れてもいいかも
CLIP has also been used in other settings of the image editing with instructions task.
\cite{TediGAN} allowed multi-modal instructions such as sketched, whereas \cite{anyface} and \cite{de-gan} assumed that a text-image paired corpus was available to refine StyleCLIP.

Similar to our study, \cite{clip2stylegan} introduced an SVM in order to utilize the normal vector for editing, although their task setting was different from ours; they restricted the instructions to pre-defined attributes.
They computed principal component vectors in the CLIP space and retrieved the positive and negative images according to the similarity to each principal component vector. 
Then, they learned the correspondence between the pre-defined attributes and the principal component vectors.
Compared with \cite{InterfaceGAN} and \cite{clip2stylegan}, we retrieve the positive and negative images with queries in the form of any natural language instruction on-the-fly.

%diffusion model
Recently, diffusion models have been used to synthesize photo-realistic images~\cite{dall-e, imagen, stablediffusion} and edit images~\cite{glide,dall-e-2,diffusionclip}. 
However, they are more computationally expensive than GANs because they generate the images with a time series model.
%faissを実行した速度を報告したい
\kosuke{
\cite{StyleGAN-T} reported that GLIDE~\cite{glide} took 10.9s to generate a 64x64 image and StyleGAN-T~\cite{StyleGAN-T} took 0.06s on an A00 GPU.
}

%The total training time was four weeks on 64 A100 GPUs using a batch size of 2048. We first trained the primary phase for 3 weeks (resolutions up to 64×64), then the secondary phase for 2 days (text embedding), and finally the primary phase again for 5 days (resolutions up to 512×512). For comparison, our total compute budget is about a quarter of Stable Diffusion’s (CompVis, 2022).
%inference speedのひかくはtable2, 3

\section{Conclusion}
%本研究は，従来研究であるStyleCLIP~\cite{StyleCLIP}を拡張し，自然言語により編集指示を与えるのみで顔画像編集が可能な手法を提案した．本研究の貢献を以下に示す．
%本研究は\cite{StyleCLIP,InterfaceGAN}とは異なり，自然言語で与えられる任意の編集指示に対してon-the-flyでCLIP空間における正例・負例集合を取得して分離超平面を導出し，得られた法線方向をStyleGANの特徴空間に射影することによって，編集前画像の説明テキスト無しに安定した顔画像編集を初めて実現した．
%顔画像編集はコンピュータビジョンにおける重要なタスクであり，メタバースにおけるアバター作成やコミュニケーション支援など産業上重要なサービスに応用できる．本技術により編集前画像の説明テキストの入力が不要となったことで，自然言語による顔画像編集技術の応用シーンを大きく拡大することができた．
\kosuke{We proposed StyleCLIP-FEU, which edits an image simply by feeding it a natural language instruction.
Different from StyleCLIP and InterfaceGAN, StyleCLIP-FEU obtains the hyperplane separating positive and negative images that are retrieved on-the-fly and uses its normal vector as the edit direction. Facial image editing is an important task with many applications, such as avatar creation and communication support. We made the image editing method easy-to-use and efficient in order to improve its practicability.}

%\newpage
{\small
\bibliographystyle{ieee_fullname}
\bibliography{reference}
}

\end{document}